# DEEP LEARNING BASED DETECTION OF FISHING VESSELS AND FISHING MONITORING USING NIGHTLIGHT IMAGES

Shantakar Mohanty[1, 2], Prasun Kumar Gupta[1], Raian Vargas Maretto[2]

[1] Indian Institute of Remote Sensing, ISRO, Dehradun, India

[2] University of Twente, Faculty of Geo-Information Science and Earth Observation (ITC), Enschede, The Netherlands

**Abstract**

The increasing demand for effective maritime surveillance has given rise to the critical need for monitoring fishing vessel activities, particularly in addressing the challenge of "dark vessels" that operate without Automatic Identification System (AIS) transmission. This study presents a novel approach for detecting small-scale fishing vessels using nighttime light (NTL) imagery from the SDGSAT-1 satellite, combined with deep learning techniques to enhance fishing monitoring awareness along the western coast of India. A dual-branch YOLO11 architecture was developed to exploit both the 10-meter panchromatic and 40-meter RGB imagery from SDGSAT-1. The custom model architecture was specifically optimized for small object detection in NTL imagery, featuring parallel convolutional backbones that process both modalities before concatenation for enhanced feature extraction. The dual-branch YOLO11 model demonstrated optimal performance with a precision of 0.99, recall of 0.93, F1-score of 0.96, and mAP@50 of 0.96, significantly outperforming single-branch implementations of YOLOv5s, YOLOv8s, and standard YOLO11s architectures. When applied to the study area covering the waters off Maharashtra, Goa, and Karnataka, the model detected 31,525 vessel instances across the temporal dataset spanning from 1st January 2022 to 31st December 2023. Cross-matching analysis with AIS data revealed that only 7,146 (22.7%) of detected vessels had corresponding AIS transmissions, while 24,379 (77.3%) were identified as potential dark vessels. Spatio-temporal analysis showed peak fishing activity during January-April, with a primary activity corridor parallel to the coastline within 50-100 km, corresponding to productive continental shelf areas. The continental shelf regions of Maharashtra, Karnataka, and Goa were identified as major fishing hotspots on the Western coast of India. This research contributes to maritime surveillance capabilities by highlighting the effectiveness of nighttime lights satellite imagery for fishing vessel detection and provides valuable insights into fishing patterns and potential regulatory compliance issues in Indian waters. The findings have important implications for fisheries management, maritime security, and sustainable ocean resource utilization.

## 1. Introduction

The monitoring and regulation of fishing activities along India's enormous coastline have posed a critical challenge in maritime fisheries and sustainable fisheries management. India's marine fisheries sector contributes approximately ₹128,011 crore to the economy, with fish and fish products generating $7.38 billion in export earnings annually (Bhatnagar, 2025; Rahman & Mehnaz, 2024). The Western Coast of India, spanning from Gujarat to Kerala for approximately 3,400 km, is one of the most significant fishing regions of the world. Illegal, unreported, and unregulated (IUU) fishing has become a critical threat to marine ecosystems and fishery industries worldwide, directly affecting sustainable fishing practices.

Traditional monitoring approaches face significant limitations in addressing these challenges. The monitoring of fishing activities is usually done using Automatic Identification Systems (AIS) and Vehicle Monitoring Systems (VMS), but AIS data is often incomplete, as vessels engaged in IUU fishing frequently disable AIS transmitters to evade detection, and several small fishing vessels do not have AIS receivers due to economic feasibility (Y. Li et al., 2023). Traditional optical remote sensing methods mostly depend on sunlight reflection and cannot provide sufficient nightlight data, which is quite crucial for vessel detection and monitoring(Zhao et al., 2023). Other conventional remote sensing methods include Synthetic Aperture Radar (SAR), which enables round-the-clock monitoring with high spatial resolution but is hampered by high noise levels and limited data widths, restricting its application for small fishing vessel detection (Song et al., 2023).

Nightlight remote sensing technologies offer a viable solution to address these monitoring limitations by providing seamless coverage and cost-effective surveillance capabilities over vast marine areas. The advancements in NTL glimmer sensors achieved a significant technological breakthrough with the launch of the SDGSAT-1 satellite in 2021, equipped with a high-resolution (10m) Glimmer Imager for Urbanization (GIU) sensor for nightlight detection. The spatial resolutions of previous sensors, DMSP-OLS (2700m), VIIRS (750m), and LJ1-01 (130m), come across as insufficient for the accurate detection and identification of small fishing vessels (Song et al., 2023).With the advent of higher resolution nightlight imagery, vessels that previously appeared as a single pixel in low-resolution sensors are now captured across multiple pixels, which can sometimes be internally disconnected, resulting in several distinct peak points within the same vessel (Zhao et al., 2023).

Deep learning has emerged as an important tool in processing NTL datasets and extracting relevant information (Jindal et al., 2024; Sahoo et al., 2020). Convolutional neural networks (CNNs) have been used for vessel detection in optical and radar satellite imagery, exhibiting high accuracy and scalability (Graziano et al., 2019). Deep learning-based approaches offer substantial potential to address these multi-pixel detection challenges through their ability to learn complex spatial patterns and contextual relationships within vessel signatures (Kamirul et al., 2025; Nie et al., 2022).

Despite the many advantages of the SDGSAT-1 and its relevance to maritime monitoring, there is a significant research gap in its applications for the same. Very few studies have attempted to systematically explore and exploit the high-resolution SDGSAT NTL data for vessel detection. This study focuses on the Arabian Sea fishing grounds on the West Coast of India, a region of great ecological and economic importance. By combining SDGSAT-1 NTL imagery, AIS data, and deep learning techniques, this study proposes a dual-branch YOLO11 model customized for fishing vessel detection. The integration of high-resolution NTL imagery with a deep learning algorithm has been used for high-accuracy object detection of fishing vessels over the study area. This study addresses not only existing gaps in maritime monitoring but also provides critical insights into the quantification of potential dark vessels based on AIS cross-matching.

## 2. Materials and Methods

### 2.1 Study Area

India is the world's second-largest producer of fish, with a major chunk of it coming from the West coast(Sarangi & Jaiganesh, 2022). The study area for this research spans the West Coast of India, which includes the coastline of the states of Maharashtra, Goa, Karnataka, and Kerala, as shown in Figure 1. This region comes under the Eastern Arabian Sea and is chosen for this study due to the high volume of fishing activities (Mandal et al., 2018) as well as maritime traffic compared to the East Coast. As per the Central Marine Fisheries Research Institute (CMFRI) report, the study area for this research witnessed 1.5 million tonnes of marine fish landings in 2023. Nighttime fishing is more prevalent on the west coast of India than on the east coast. This is due to the broader continental shelf and higher biological productivity of the Arabian Sea compared to the Bay of Bengal (Sarangi & Jaiganesh, 2022). A 200 km buffer zone into the Arabian Sea encompassing a total area of 3,18,208 sq. km has been created along the west coast of India for fishing vessel detections using SDGSAT-1 NTL and AIS data.

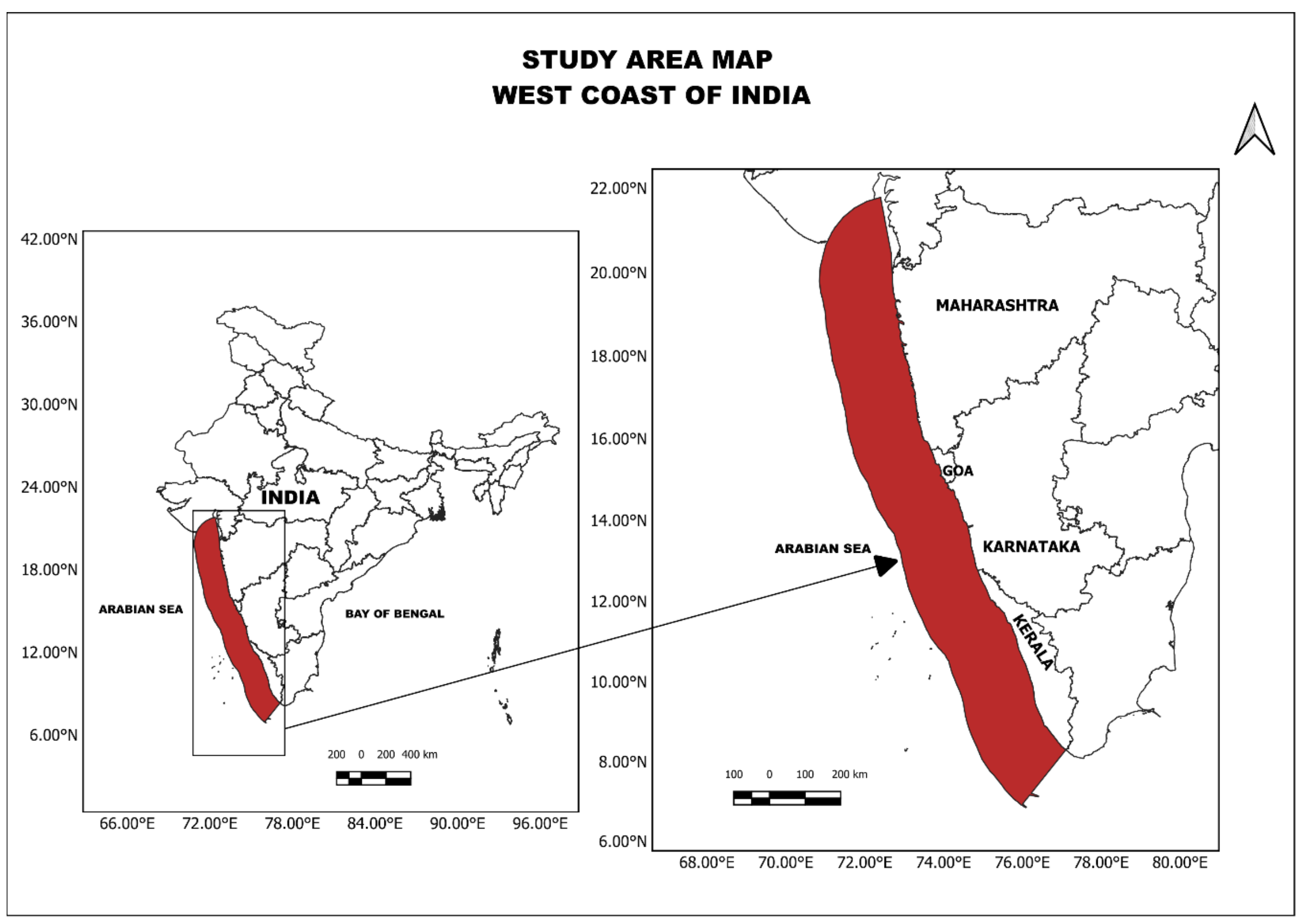

**Figure 1 Study Area Map**

### 2.2 Datasets and Pre-processing

#### 2.2.1 SDGSAT-1 Glimmer Imager for Urbanization (GIU) Images

The SDGSAT-1 mission, launched in 2021, is the first scientific satellite dedicated to the Sustainable Development Goals (SDGs) (Guo et al., 2023). SDGSAT-1 carries three advanced sensors for accurate detection and analysis of human activity signals: a thermal infrared spectrometer, a glimmer imager, and a multi-spectral imager.

The SDGSAT-1 Glimmer Imager for Urbanization (GIU) provides high-resolution nighttime light (NTL) imagery that offers considerable advancements in detecting artificial light sources from fishing vessels. The GIU sensor has a spatial resolution of 10 m for panchromatic and 40 m for RGB bands. This enhanced resolution allows the GIU sensor to capture detailed nighttime light data, allowing for more precise observations of fishing activities across the study area. It further covers a spectral range of 450-900 nm for panchromatic and specific ranges for RGB bands, as seen in Table 1, enabling multispectral analysis of nighttime lights. For this study, 168 images have been acquired over two years (January 2022 to December 2023) and then pre-processed and fed into a lightweight YOLO11 model for the object detection of fishing vessels.

| Type | Index | Specifications |
|---|---|---|
| Orbit | Type | Sun-synchronous Orbit |
| | Altitude | 505 km |
| | Inclination | 97.5° |
| | Revisit Time | 11 Days |
| GIU | Swath Width | 300 km |
| | Glimmer Imager Bands | P: 444-910 nm<br>B: 424-526 nm<br>G: 506-612 nm<br>R: 600-894 nm |
| | Spatial Resolution of Glimmer Imager | P: 10 m, RGB: 40 m |
| | Overpass time | 16:30 UTC, 22:00 Local time |

**Table 1 SDGSAT Specifications**

Source - https://www.sdgsat.ac.cn/satellite/describe

### 2.2.2 Automatic Identification System (AIS) Data

AIS (Automatic Identification System) is a vessel-tracking technology that transmits real-time information such as vessel identification, position (latitude and longitude), speed, heading, and navigational status via VHF radio signals. Originally mandated for safety and collision avoidance, AIS has become a critical data source in maritime monitoring, particularly for identifying fishing activity, vessel behaviour, and spatiotemporal movement patterns.

For this study, AIS data have been acquired from Global Fishing Watch (GFW), a non-profit organization that processes and provides open-access data on fishing activity worldwide. GFW leverages machine learning algorithms, vessel registry databases, and manual expert review to identify fishing vessels from the vast amount of AIS data.

To ensure temporal synchronization with the satellite imagery, the AIS dataset was filtered to retain only those records occurring within a ±30-minute window of the SDGSAT-1 overpass time. This strict temporal constraint is crucial for minimizing positional errors caused by vessel displacement between the image capture and the AIS transmission.

### 2.2.3 Image Patching and Land Masking

With a swath of 300 kms raw SDGSAT-1 scenes are too large for direct ingestion into deep learning models due to GPU memory constraints. A patching strategy was implemented:

1) **Tiling:** The high-resolution Panchromatic images were tiled into 256 x 256 pixel patches. To maintain geospatial alignment, the corresponding lower-resolution RGB images were tiled into 64 x 64 pixel patches. This 4:1 ratio corresponds exactly to the 10m:40m resolution difference.

2) **Land Masking:** Since the objective is maritime surveillance, terrestrial light sources (cities, coastal highways) constitute false positives. A high-resolution shapefile of the Indian coastline was used to mask land areas. Patches containing > 80% land were discarded. Patches with partial land cover (<80%) were retained to ensure the detection of nearshore fishing activity, which is often dense in the shallow continental shelf waters.

### 2.2.4 Radiometric Normalization and Enhancement

NTL imagery is characterized by a high dynamic range and sparse data distribution (mostly dark pixels). Raw digital number (DN) values can vary significantly based on gain settings and lunar illumination. To stabilize model training, a normalization procedure was applied:

$$P_{norm} = \frac{P_{original} - P_{min}}{P_{max} - P_{min}} \times 255$$

$P_{norm}$ - Normalized pixel value after scaling.
$P_{original}$ - Original pixel value before normalization.
$P_{min}$ - Minimum pixel value of the image
$P_{max}$ - Maximum pixel value of the image

This linear scaling to an 8-bit integer range acts as a contrast enhancement technique, amplifying faint light signatures against the dark ocean background.

## 2.3 Deep Learning Architecture Design

Following a thorough literature analysis(Hu et al., 2024; Khanam & Hussain, 2024; Shao et al., 2021; Song et al., 2023) and model comparisons, this study adopted Ultralytics YOLO11, the latest advancement in the YOLO (You Only Look Once) series that delivers state-of-the-art real-time object detection capabilities with enhanced computational efficiency.

Performance evaluations across benchmark datasets demonstrate that YOLO11 achieves higher mean Average Precision (mAP) than its predecessors (YOLOv8, v10), with fewer parameters and lower latency, particularly in lightweight configurations (e.g., YOLOv11-n/s/m)(Khanam & Hussain, 2024).

Given the nature of SDGSAT-1 NTL imagery, characterized by small, faint fishing vessel signals and fragmented light clusters, a customized (more compact) version of YOLO11 was created by modifying the original architecture. This version streamlines the architecture (removing unnecessary detection heads and channels) to focus on small-object detection and low-frequency imagery, reduces the chances of overfitting with limited training samples, and accelerates inference on dual-branch imagery inputs.

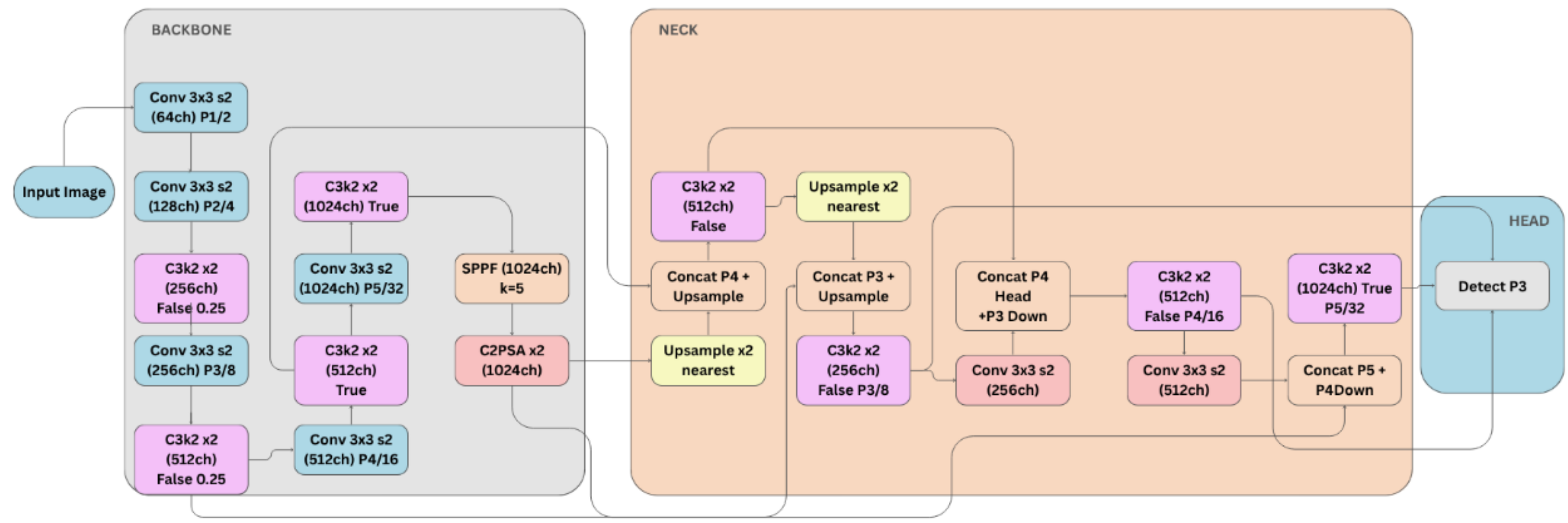

**Figure 2 YOLO 11 Architecture**

### 2.3.1 Model Configuration of Dual-Branch YOLO11

The original YOLO11 architecture, as seen in **Figure 2**, is designed to process a single input modality (usually RGB). In contrast, SDGSAT-1 offers dual imagery in the form of:

- Panchromatic (LH) bands at 10m spatial resolution, and
- RGB imagery at a coarser 40m spatial resolution.

To exploit both the 10 m panchromatic (LH) and 40 m RGB glimmer imagery, a dual-branch YOLO11 architecture was developed. This design features two parallel convolutional backbones: one processes LH inputs and the other processes RGB, as illustrated in **Figure 3**.

The dual-branch YOLO11 architecture deviates from the conventional single-input approach, incorporating specialized dual-input branches to optimize processing of multispectral SDGSAT-1 imagery. While the original YOLO11 model handles only a single 3-channel RGB input, the modified version employs two separate convolutional streams designed to exploit the complementary characteristics of panchromatic and RGB data. The LH (panchromatic) branch processes high-resolution 256×256 panchromatic patches through dedicated convolutional layers that progressively downsample the input to 64×64 resolution, preserving critical fine-scale spatial information essential for small vessel detection. Simultaneously, the RGB branch accepts inputs that are pre-resized to 64×64 resolution to ensure spatial compatibility during the subsequent fusion process, eliminating resolution mismatches that could compromise feature integration effectiveness.

The architecture implements early fusion at 64×64 resolution (40m), representing a critical design decision that balances spatial detail preservation with computational efficiency. The two branches undergo concatenation after initial feature extraction at 40m resolution, which ensures spatial compatibility while enabling the exploitation of both panchromatic and RGB image characteristics. This early fusion strategy allows the model to leverage the high spatial resolution advantages of panchromatic imagery while incorporating the spectral information provided by RGB channels, creating a unified feature representation that captures both fine spatial details necessary for small vessel detection and spectral characteristics that distinguish vessels from background noise.

The model employs a deliberately shallow backbone architecture by removing deeper hierarchical layers, including C3k2, SPPF, and C2PSA components, from the original YOLO11 backbone. This architectural simplification represents a deliberate design choice customized to the specific characteristics of the vessel detection task, recognizing that the relatively simple object class (small illuminated vessels) does not require the complex feature hierarchies needed for multi-class natural image detection. The shallow backbone design prioritizes the retention of fine spatial cues that are critical for small object detection while reducing computational expense and minimizing the risk of over-smoothing that could eliminate subtle vessel signatures in nighttime imagery.

The architecture incorporates a compact detection head that performs inference at only two scale levels (P3 and P4) rather than the traditional multi-scale detection at P3, P4, and P5 levels employed in the original YOLO11 framework. This scale reduction strategy reduces computational cost while specifically addressing the challenge of false positive detections from coarse-resolution layers, which often misfire on glimmer noise and other artifacts common in nighttime maritime imagery.

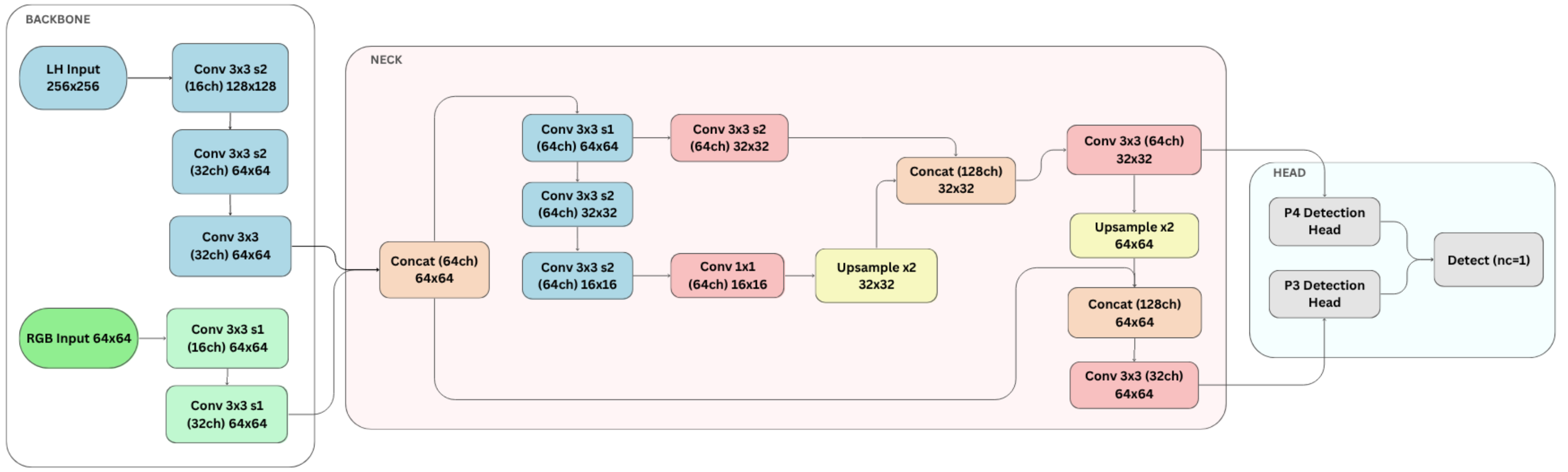

**Figure 3 Light Dual-branch YOLO11 Architecture**

## 2.4 Model Training and Evaluation

Following hyperparameter optimization, the dual-branch YOLO11 model was trained on the complete training dataset using the identified optimal configuration. The training protocol incorporated the following parameters: 200 epochs with a batch size of 16, input image dimensions of 256 × 256 pixels, and an early stopping mechanism with 50-epoch patience to prevent overfitting. Validation was conducted after each epoch to continuously monitor model performance and convergence behavior. This structured training pipeline enabled the model to effectively capture the spatial-spectral signatures of lit fishing vessels across varied oceanic backgrounds and noise conditions while maintaining robust generalization to unseen data.

Model performance was assessed using standard object detection metrics alongside training-specific loss functions. The evaluation framework comprised three primary loss components monitored throughout training: (1) box loss, quantifying localization accuracy through advanced IoU-based formulations such as Complete IoU (CIoU) or Efficient IoU (EIoU) that account for bounding box overlap, center distance, and aspect ratio; (2) classification loss, calculated using Binary Cross-Entropy (BCE) to penalize misclassifications in the fishing versus non-fishing vessel categorization; and (3) Distribution Focal Loss (DFL), which models bounding box regression as a probability distribution over discrete bins to enhance coordinate prediction precision, particularly critical for small vessel detection.

Detection performance was quantified through four key metrics. Precision measures the proportion of correctly identified fishing vessels among all detections, with higher values indicating fewer false positives. Recall captures the proportion of actual fishing vessels successfully detected, where elevated scores signify minimized false negatives. Mean Average Precision (mAP) was computed at two strictness levels: mAP@50, representing average precision at IoU threshold of 0.5, and mAP@50-95, averaged across ten IoU thresholds (0.5 to 0.95 in 0.05 increments) to provide evaluation of both localization and classification capabilities. The F1 score, computed as the harmonic mean of precision and recall, offered a balanced performance summary, particularly valuable when trading off false positives against false negatives. These metrics were calculated at each epoch using the validation set, enabling real-time assessment of learning dynamics and informing decisions regarding early stopping and hyperparameter refinement based on loss convergence patterns and metric progression.

## 2.5 AIS Cross-Matching Framework

To validate SDGSAT-1 vessel detections and identify potential dark vessel activity, raw AIS data were first temporally filtered to a ±30-minute window around satellite overpass times and spatially converted to 0.01° × 0.01° grid cells before implementing a systematic spatial-temporal cross-matching procedure.

### 2.5.1 Buffer Radius Determination Based on Vessel Speed Analysis

The spatial buffer radius for cross-matching was derived from empirical vessel speed data specific to Arabian Sea fishing operations. A comprehensive literature review identified operational speeds across multiple vessel types and fishing methods along India's west coast: trawlers during towing operations (3–6 knots) and steaming (7–8 knots), ring-seine vessels (5.5–7.5 knots), deep-sea long-liners (7–8 knots), and motorized canoes (5–7 knots) (Andersen, 1992; *Annual Report ICAR-CIFT 2021*; Dineshbabu, 2013; Singh et al., 2023). An average speed of 7 knots was adopted as representative of typical fishing vessel operations, balancing lower active fishing speeds against higher transit velocities.

Considering the ±30-minute temporal matching window between satellite observation and AIS recording, the buffer radius was calculated as:

$$\text{Buffer Radius} = 7 \text{ nautical miles/hr} \times 0.5 \text{ hours} \times 1.852 \text{ km/nautical mile} = 6.5 \text{ km}$$

Each detection polygon was buffered using a 6.5 km radius, creating a spatial uncertainty zone that accounts for potential vessel displacement during the temporal window. These buffered areas were intersected with AIS vessel polygons using a two-stage filtering approach: spatial overlap followed by temporal constraint validation. A match was considered valid only if the detection date and time coincided with either the AIS entry or exit timestamp, ensuring that only vessels present in the vicinity at the time of satellite observation were retained as valid matches.

### 2.5.2 Stepwise Cross-Matching Procedure

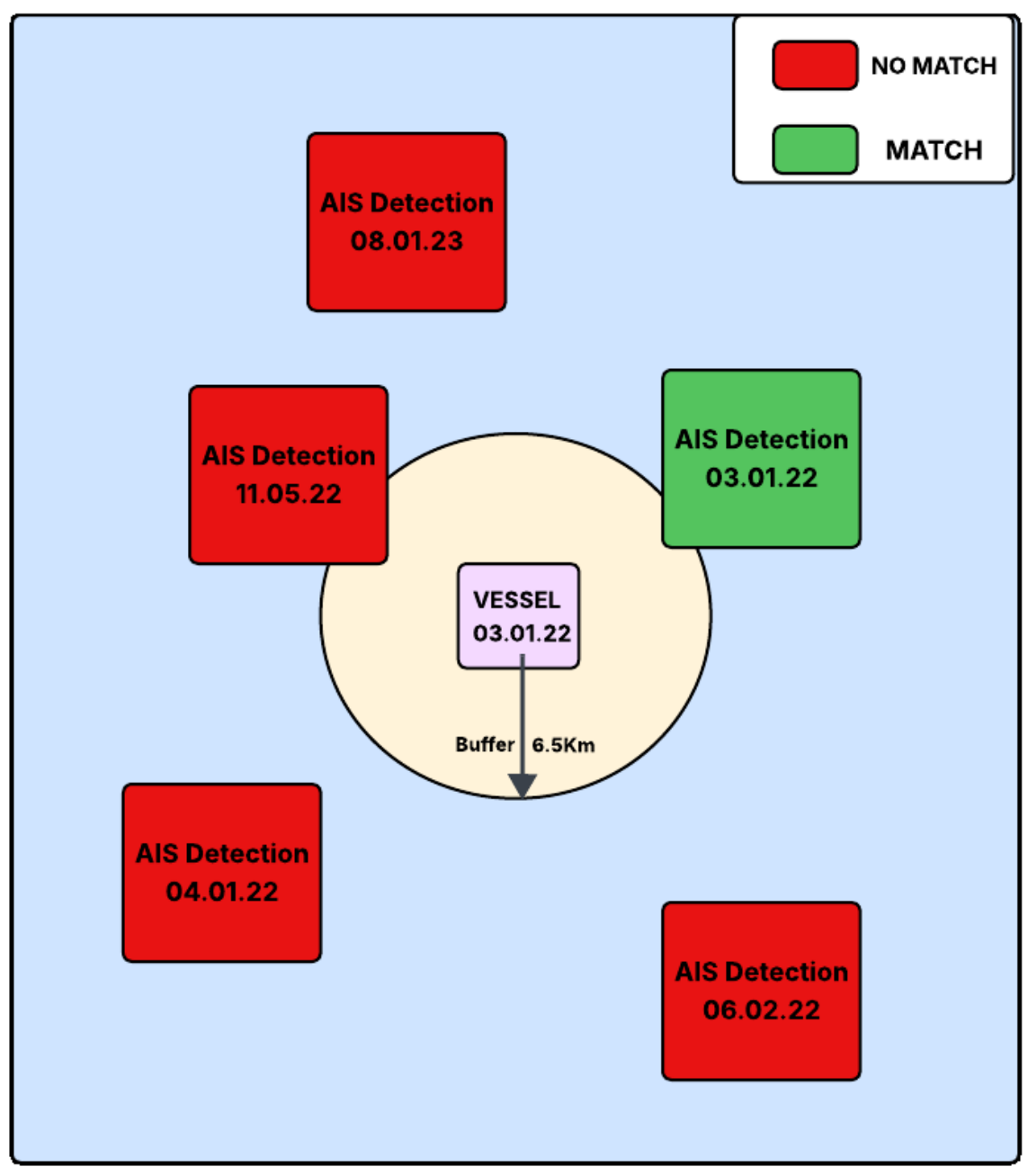

**Figure 4 AIS cross-matching logic**

The cross-matching procedure follows a systematic three-stage approach designed to ensure robust spatial-temporal correlation between SDGSAT-1 detections and AIS records, as illustrated in **Figure 4**.

**Stage 1: Data Preprocessing and Standardization**

Both SDGSAT-1 vessel detections and AIS datasets were converted to standardized GeoJSON format. Detection polygons were extracted from YOLO11 bounding boxes and transformed to geographic coordinates, while AIS point data were converted to 0.01° × 0.01° grid cells to approximate spatial uncertainty and maintain compatibility with Global Fishing Watch datasets.

**Stage 2: Spatial Buffer Generation and Intersection**

Each SDGSAT-1 detection polygon was buffered by a 6.5 km radius, creating a circular uncertainty zone around each detected vessel **(Figure 4).** AIS grid cells falling within any detection buffer zone were identified using GeoPandas spatial intersection operations, reducing computational complexity by eliminating geographically distant AIS records.

**Stage 3: Temporal Constraint Application**

For each spatially overlapping pair, strict temporal matching criteria were applied. SDGSAT-1 detection timestamps, derived from image acquisition metadata, were compared against AIS temporal windows, with records considered valid matches only if their timestamps aligned with the detection date and time. When multiple AIS polygons satisfied both spatial intersection and temporal overlap criteria for a single detection, temporal constraints served as the primary filtering mechanism, retaining only records whose timestamps matched the detection date. In cases where multiple AIS polygons remained after temporal filtering, a proximity-based selection algorithm calculated Euclidean distances between detection and AIS polygon centroids, retaining only the nearest AIS record as the final match.

## 2.6 Spatio-Temporal Analysis

To understand fishing activity patterns across the study area, a comprehensive spatio-temporal analysis was conducted using model-derived vessel detections to uncover trends in spatial distribution and temporal variability at multiple scales.

### 2.6.1 Temporal Analysis Framework

Temporal dynamics were examined across multiple time scales to identify cyclical fishing behaviors, seasonal patterns, and long-term trends. Monthly analysis visualized vessel detection counts through time series line plots across the entire study period and bar charts showing average monthly patterns to reveal seasonal trends and peak activity periods. Quarterly analysis aggregated data into four quarters (Q1: Jan-Mar, Q2: Apr-Jun, Q3: Jul-Sep, Q4: Oct-Dec) to identify broader seasonal patterns while maintaining sufficient temporal resolution. Day-of-year analysis examined vessel activity across all 365 days using scatter plots with 7-day rolling averages for trend smoothing, incorporating month boundary indicators and statistical trend analysis to identify fine-scale temporal patterns and recurring hotspots throughout the maritime calendar year.

### 2.6.2 Spatial Analysis Framework

Spatial analysis employed hexagonal binning (hexbin) methodology to transform discrete vessel detections into density heatmaps, offering significant advantages over traditional point plotting by reducing overplotting, enabling visual aggregation of nearby detections, and providing statistical representation of spatial clustering patterns. Hexbin parameters were optimized with gridsize values ranging from 25 (monthly subplots) to 100 (overall heatmaps) to balance spatial resolution with visual clarity. Monthly spatial heatmaps were generated by combining data from all available years to reveal seasonal spatial shifts and month-specific hotspots, while a comprehensive heatmap combining all detections across the entire study period provided the master spatial distribution pattern and identified consistently active maritime areas.

### 2.6.3 Advanced Analysis for Cross-Matched and Dark Vessels

Differentiation between cross-matched vessels (with corresponding AIS records) and dark vessels (without AIS transmission) required specialized analytical approaches. AIS cross-match coverage percentage analysis quantified surveillance effectiveness within hexagonal grid cells using:

$$\text{Coverage}_{\%} = \frac{\text{AIS matched count}}{\text{Total count}} \times 100$$

producing values from 0% (complete absence of AIS matches) to 100% (exclusive AIS presence). Spatial dominance analysis quantified the relative prevalence of dark versus AIS-matched vessels by computing dominance indices:

$$\text{Dominance} = \frac{\text{Dark count} - \text{AIS matched count}}{\text{Dark count} + \text{AIS matched count}}$$

identifying geographic zones with concentrated potential dark vessel activity. Coverage gap analysis identified areas with high dark vessel activity but low AIS coverage through separate normalization of dark and AIS vessel counts:

$$\text{Normalized}_{\text{dark}} = \frac{\text{Dark count} - \min_{\text{dark}}}{\max_{\text{dark}} - \min_{\text{dark}}}$$

$$\text{Normalized}_{\text{AIS}} = \frac{\text{AIS count} - \min_{\text{AIS}}}{\max_{\text{AIS}} - \min_{\text{AIS}}}$$

followed by gap index calculation as $\text{Gap Index} = \max(\text{Normalized}_{\text{dark}} - \text{Normalized}_{\text{AIS}}, 0)$, with red-scaled heatmaps highlighting significant monitoring blind spots.

## 3. Results

The dual-branch YOLO11 model demonstrated stable convergence across all loss components during the 200-epoch training period. The training process exhibits positive learning patterns, indicating successful model optimization and generalization. The validation scores of the model are presented in **Table 2**.

| Metrics | Score |
|---|---|
| **Precision** | 0.972 |
| **Recall** | 0.888 |
| **mAP50** | 0.943 |
| **mAP50-95** | 0.494 |
| **F1 Score** | 0.928 |

**Table 2 Validation Metrics**

### 3.1 Model Training and Convergence

The Dual-Branch YOLO11 model was trained for 200 epochs with a batch size of 16 using the AdamW optimizer. The training process exhibited robust convergence characteristics. As illustrated in the training logs **(Figure 5)**, the Box Loss decreased rapidly from an initial value of ~4.5 to stabilize around 1.8 within the first 25 epochs. This rapid learning curve suggests that the geometric features of vessels in NTL imagery are distinct and learnable.

Crucially, the validation loss closely tracked the training loss throughout the process, with no sign of divergence. This indicates that the model generalized well and did not overfit to the training set, a common risk when working with relatively small datasets of specialized imagery. The Distribution Focal Loss (DFL) stabilized around 1.0, confirming the model's ability to learn the uncertainty distributions of the bounding box edges.

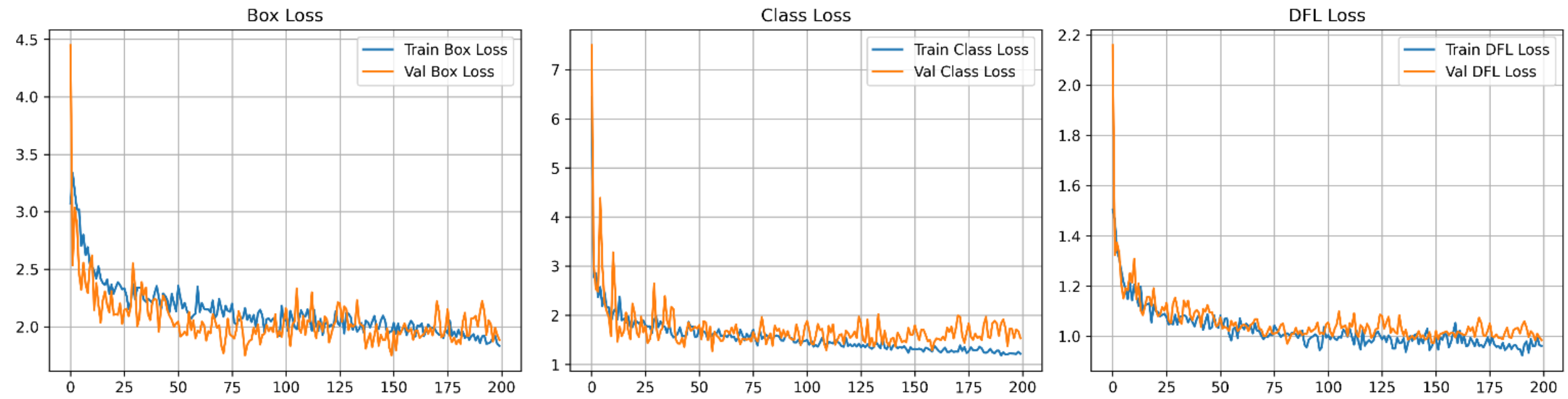

**Figure 5 Loss Curves**

## 3.2 Detection Performance Assessment

The model's performance was evaluated on the independent test set using standard object detection metrics, as seen in **Table 3**. The Dual-Branch YOLO11 achieved suitable results, demonstrating its suitability for operational surveillance.

| Metrics | Score |
|---|---|
| **Precision** | 0.994 |
| **Recall** | 0.934 |
| **mAP50** | 0.966 |
| **mAP50-95** | 0.457 |
| **F1 Score** | 0.965 |

**Table 3 Test Metrics**

## 3.3 Model Detection Results

Following the completion of model training and test performance evaluation, the final YOLO11 dual-branch architecture optimized through hyperparameter tuning was deployed for large-scale inference over SDGSAT-1 GIU imagery. The objective of this phase was to operationalize the trained model to identify fishing vessels across vast temporal and spatial extents of the study area and to extract geolocated detections for further cross-matching and analysis.

A total of over 1.5 million image patches were extracted from pre-processed SDGSAT-1 GIU scenes spanning the years 2021 and 2022.

Out of the 1.5 million patches, a total of 31,526 potential fishing vessels were detected across all processed scenes, as shown in **Figure 11**. The average confidence score for these detections ranged between 0.4 and 0.7, indicating moderate to high confidence across most predictions.

Below are some of the detection results across several scenarios and noise conditions:

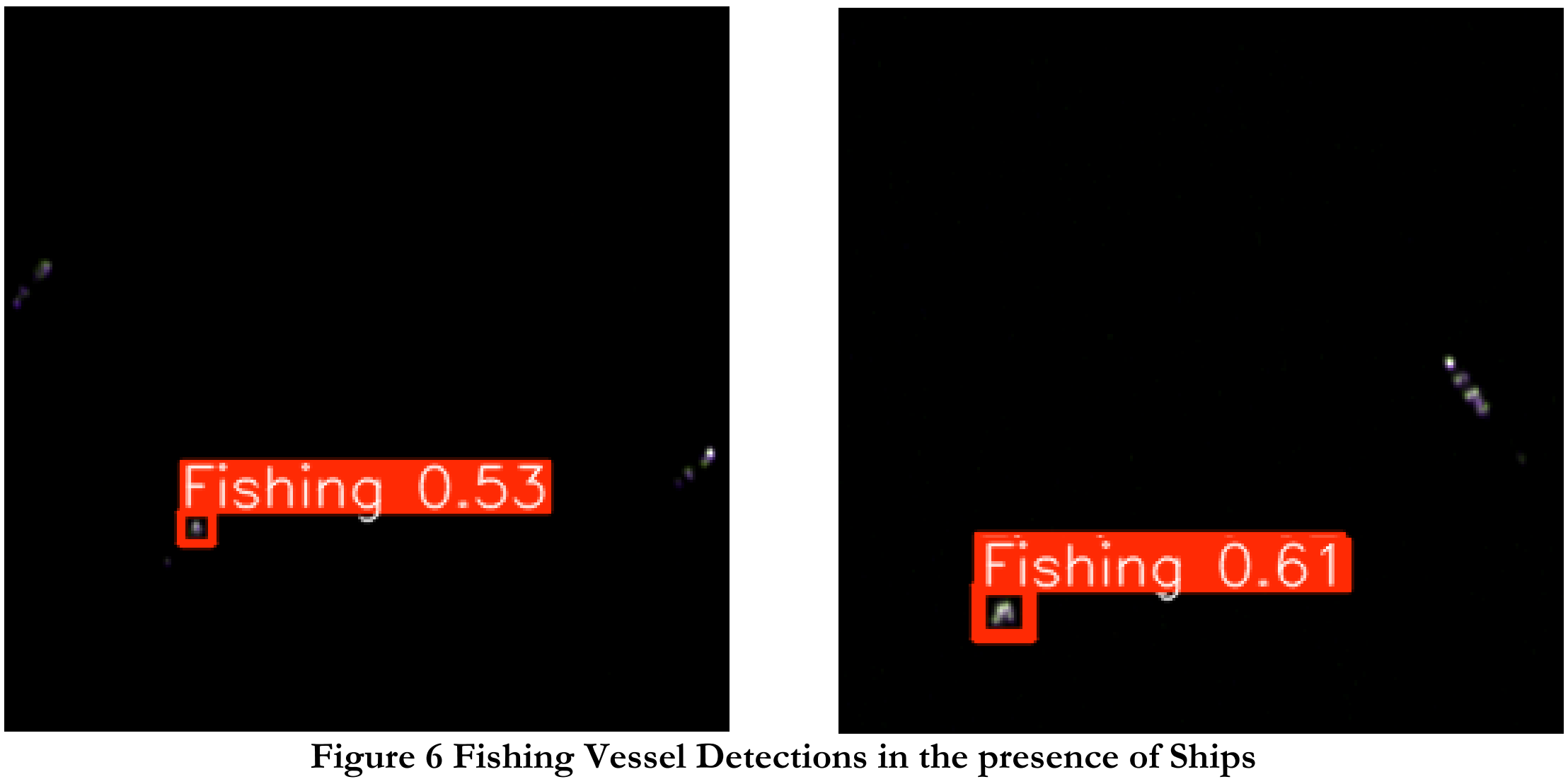

Figure 6 Fishing Vessel Detections in the presence of Ships

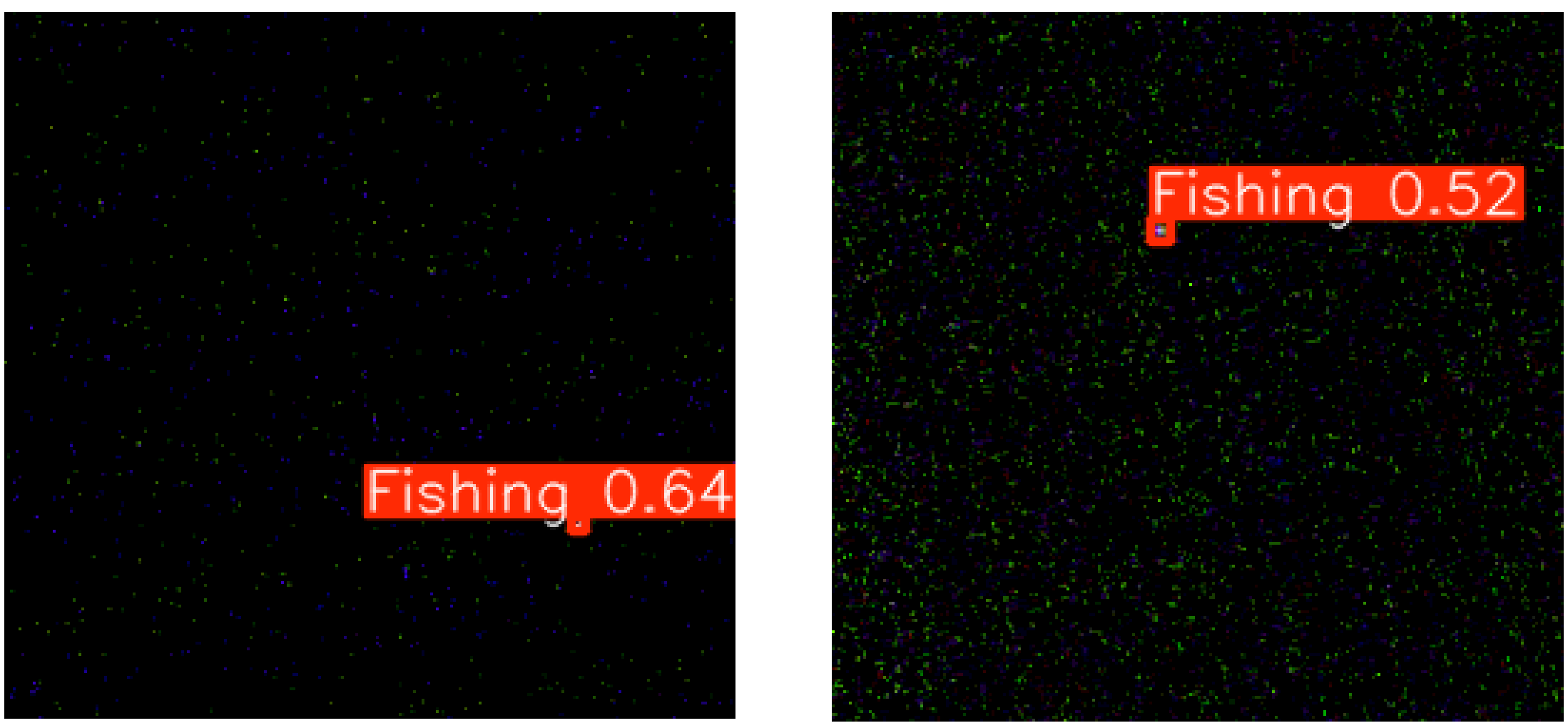

Figure 7 Fishing Vessel Detections in Extremely Noisy Conditions

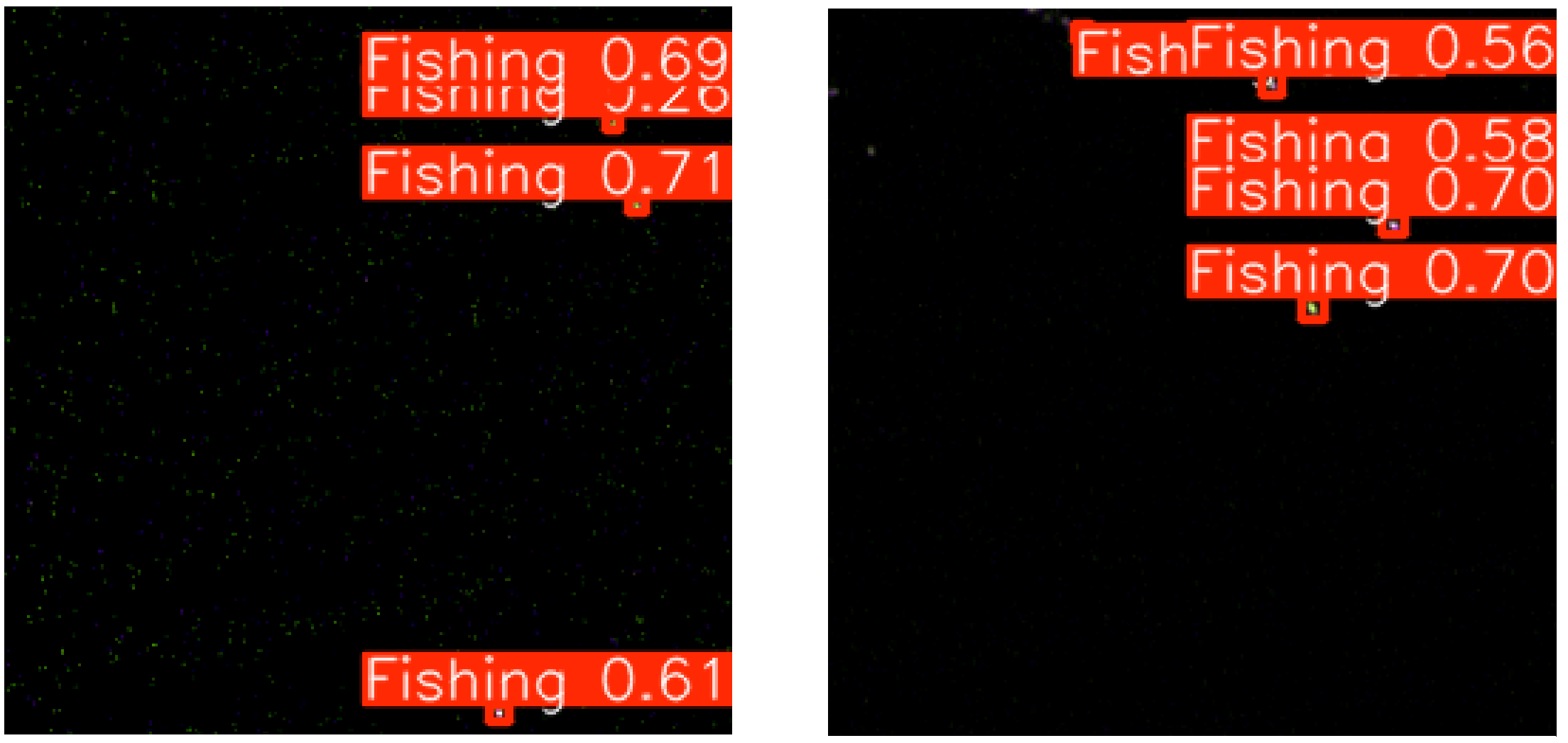

Figure 8 Multiple Vessel Detections in a single patch

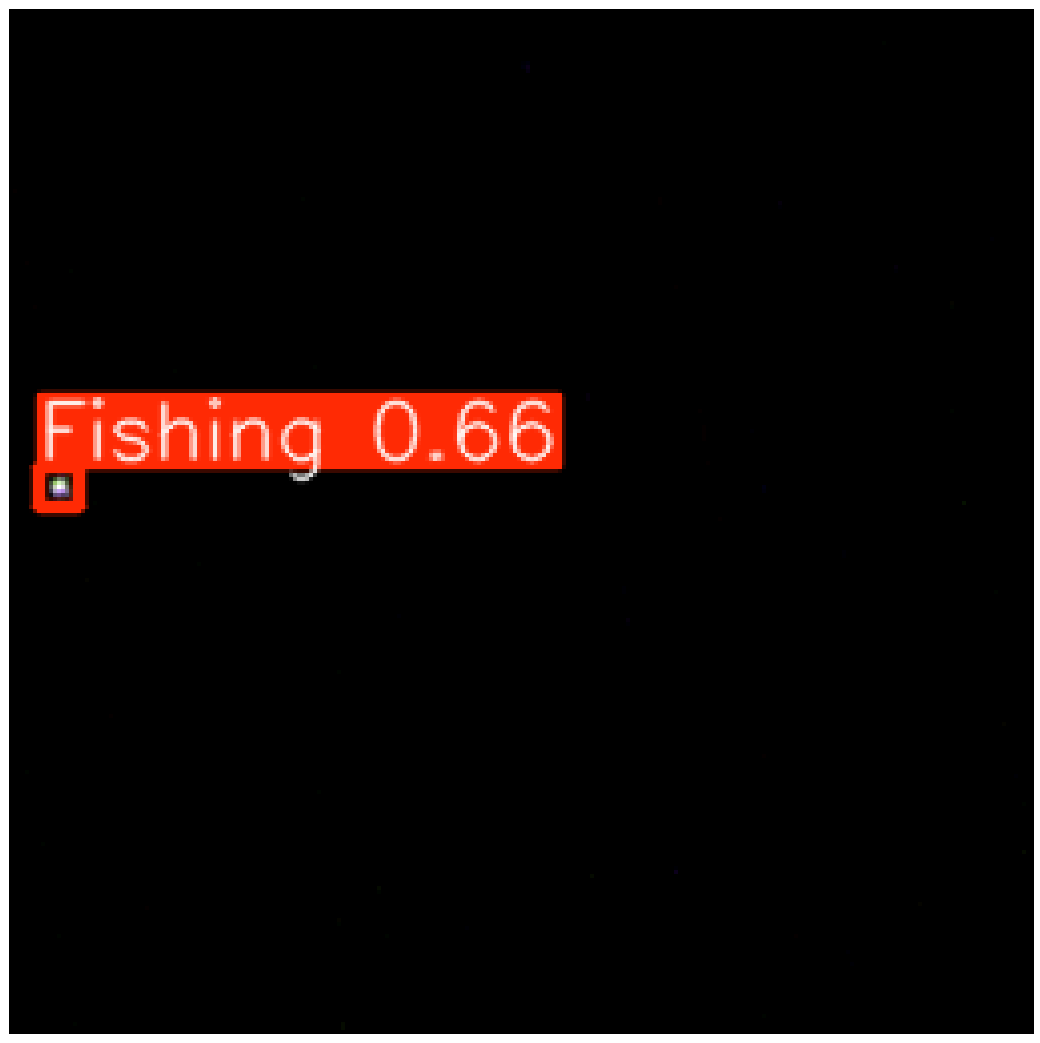

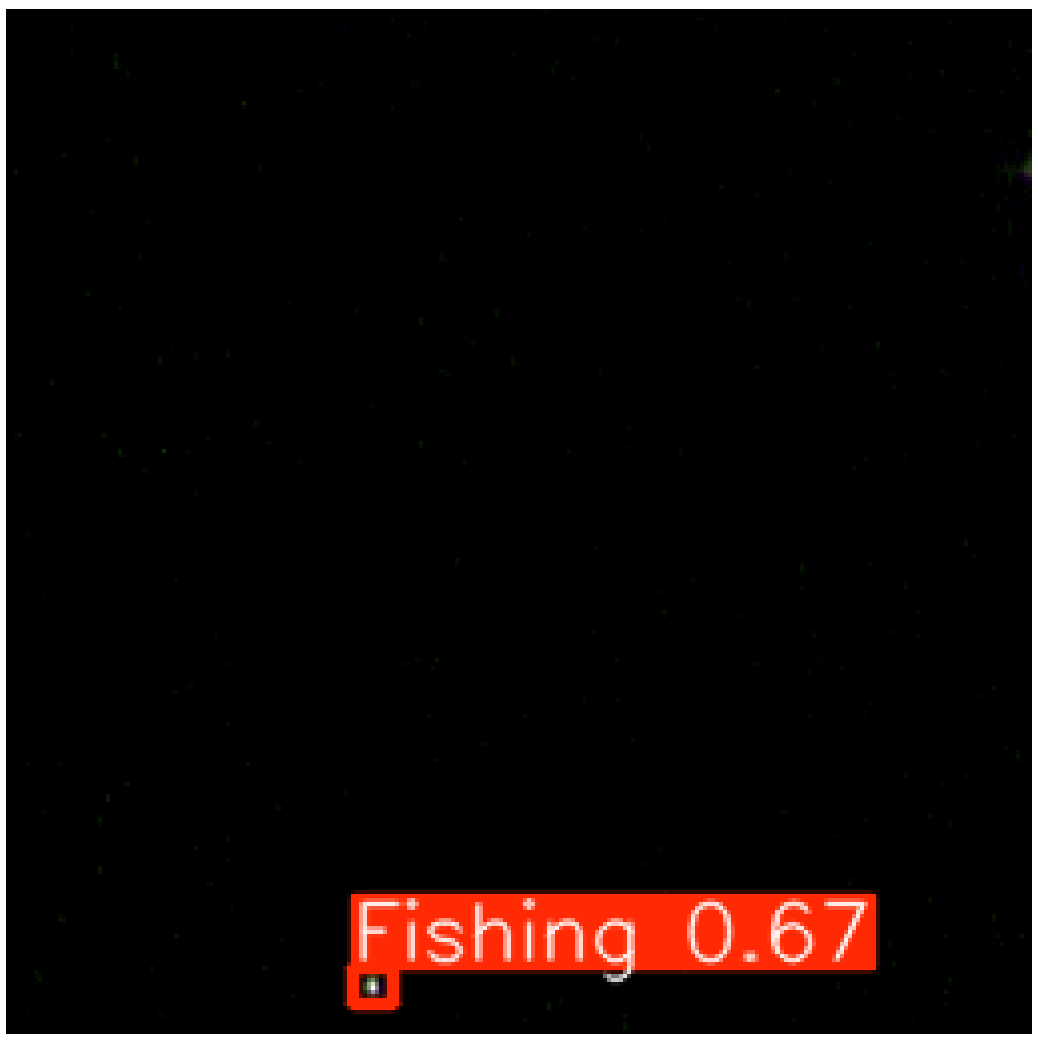

**Figure 9 Detections in clean/no noise conditions**

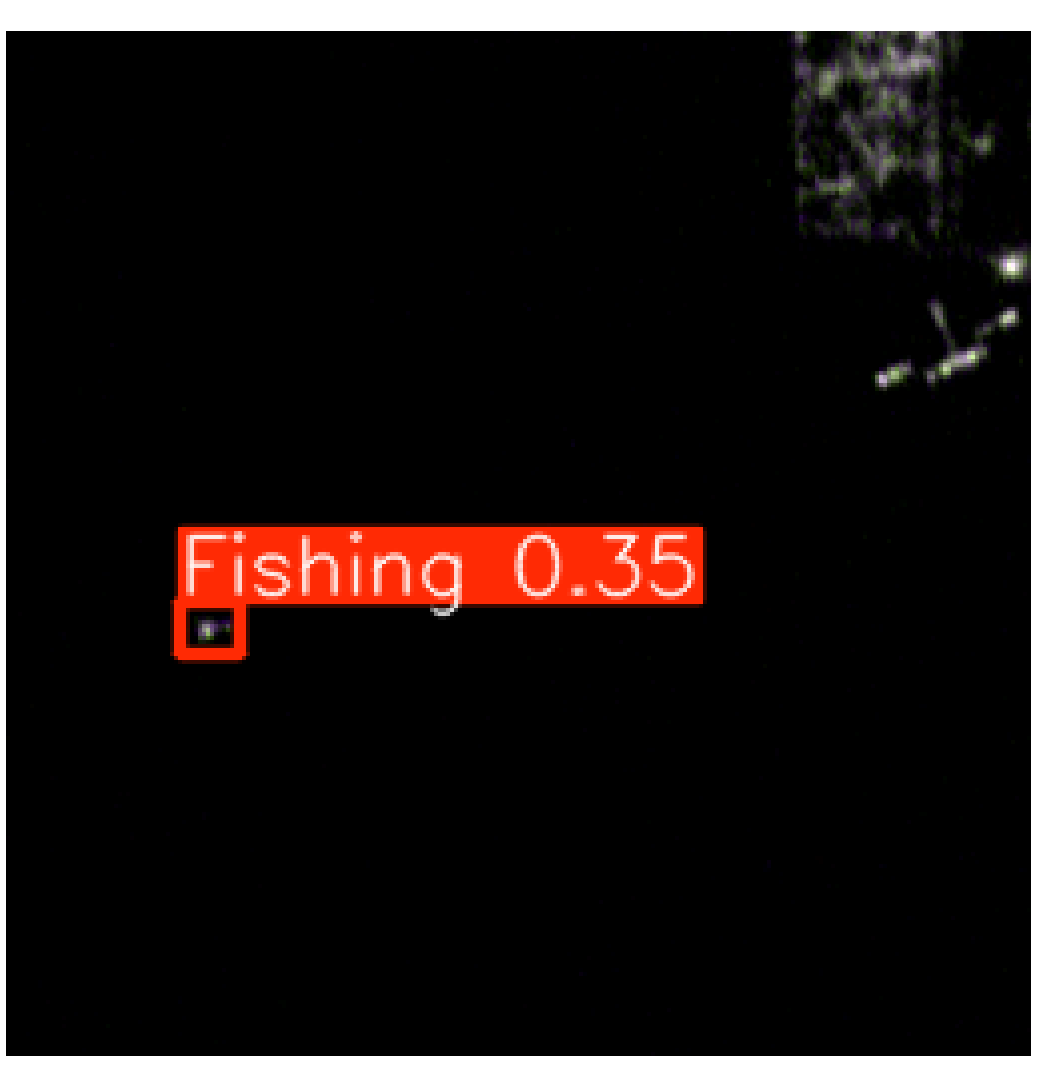

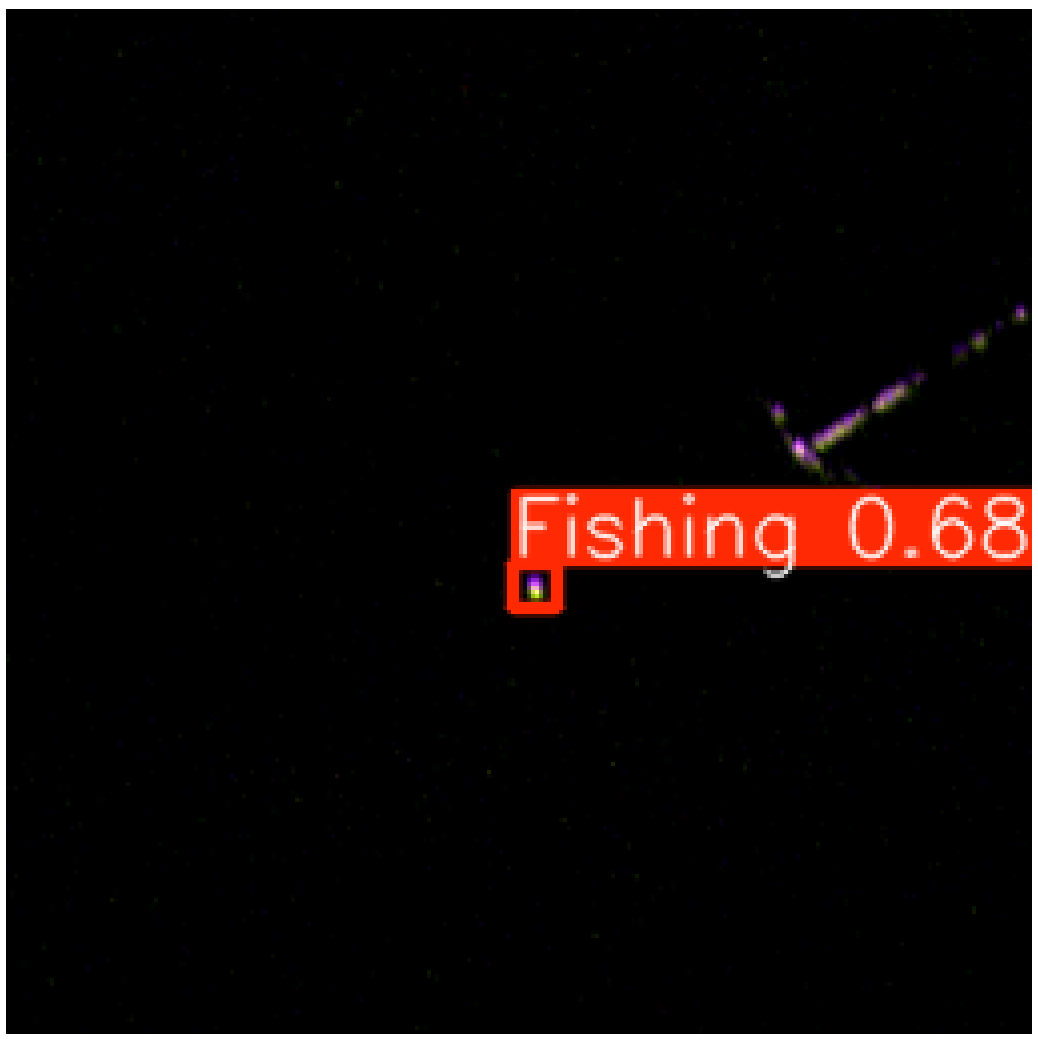

**Figure 10 Vessel Detections in near-shore conditions**

The dual-branch YOLO11 model demonstrates robust detection capabilities across diverse maritime environments and operational scenarios. The algorithm effectively identifies fishing vessels in complex environments with large commercial vessels, maintaining accurate discrimination between fishing vessels and ships (**Figure 6**). Under challenging conditions characterized by significant noise, the detection system continues to perform reliably, demonstrating resilience to suboptimal imaging conditions (**Figure 7**). The model exhibits strong scalability in high-density fishing areas, successfully detecting multiple vessels within a single image patch while maintaining individual vessel identification accuracy (**Figure 8**). Under optimal conditions with minimal noise interference, the detection algorithm achieves high precision, providing clear delineation of vessel boundaries and confident classifications (**Figure 9**). The system maintains consistent performance in near-shore environments where complex coastal features and potential interference from terrestrial illumination sources could compromise detection accuracy ( **Figure 10**).

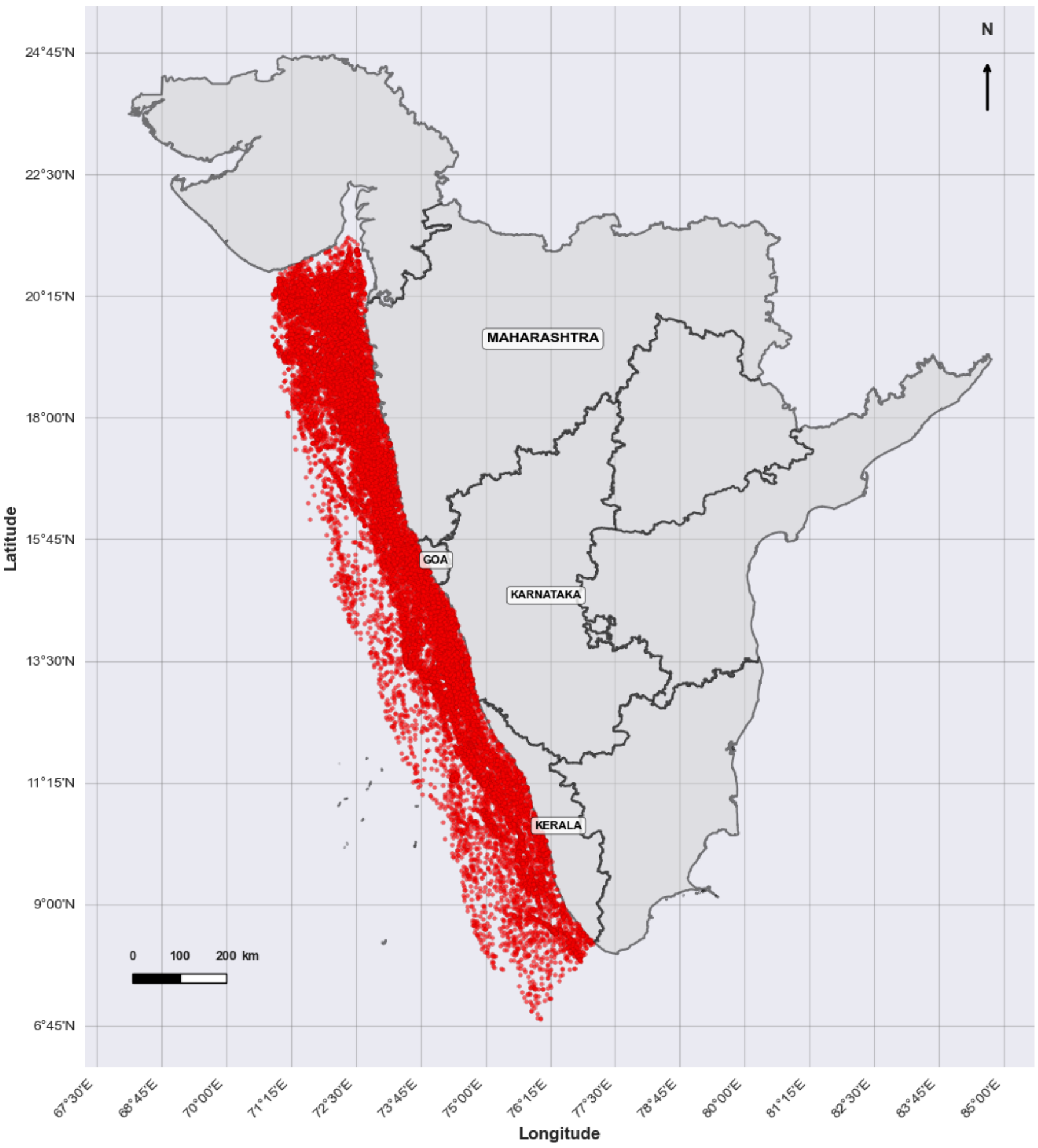

**Figure 11 All Vessel Detections 2022-23**

## 3.4 AIS Cross-Matching Results

The results of the spatio-temporal cross-matching process are shown in **Table 4**. With a match rate of 22.7%, the results indicate a substantial presence of vessels operating without active AIS transmission, with over 77% of all detections falling into this category of dark vessels. While it is important to acknowledge that not every dark vessel is engaged in illegal activity, the lack of AIS signals is concerning, especially given that Indian maritime regulations require all vessels over 20 meters in length to carry and actively transmit AIS data (*Hidden Tides*, 2025).

| Metric | Count | Percentage |
|---|---|---|
| **Total Vessel Detections** | 31,525 | 100% |
| **AIS-Matched Vessels** | 7,146 | 22.7% |
| **Potential Dark Vessels** | 24,379 | 77.3% |

**Table 4 AIS Match Rates**

The spatial distribution of dark vessels shows a dense concentration along the entire western coastline of India, forming a distinctive corridor from 24°N to 6°N latitude. The highest density occurs in nearshore waters between 15°N and 20°N. This suggests a prevalence of smaller fishing vessels (potentially below the 20m AIS mandate threshold) or significant non-compliance in these productive zones. In contrast, AIS-matched vessels exhibited a more dispersed distribution, extending further offshore into deeper waters, typical of larger commercial fleets

## 3.5 Spatio-temporal Distribution of Detected Vessels

**Temporal Patterns:** The analysis revealed pronounced seasonality as seen in **Figure 12**. Peak fishing activity occurs from January through April (Q1), accounting for 62.7% of all annual detections (20,554 vessels). Activity collapses during the monsoon months (July-September), with Q3 recording only 79 detections (0.24%), reflecting the impact of severe weather and the annual fishing ban. Activity recovers in Q4 (October-December) with 12.5% of annual detections.

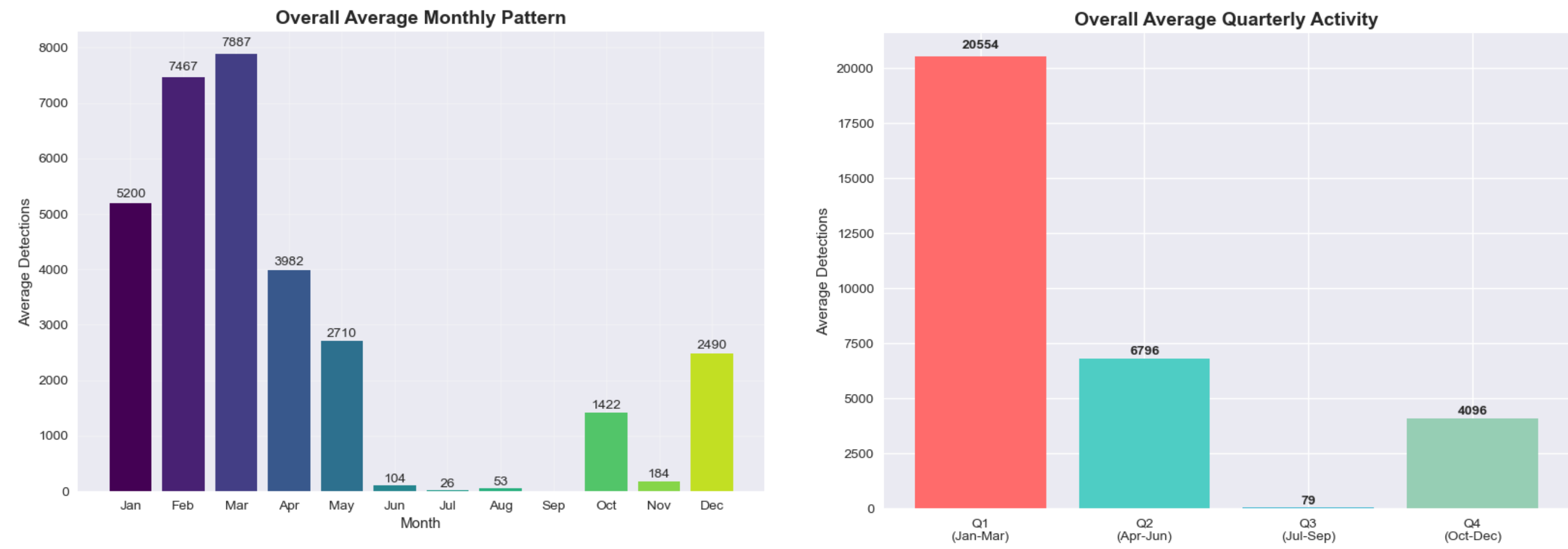

**Figure 12 Temporal Trends**

### (i) AIS Cross-match Coverage Percentage (Local Analysis)

The AIS cross-match coverage percentage analysis showcased in **Figure 13** provides the foundational spatial assessment of surveillance effectiveness by quantifying the proportion of detected vessels that successfully correlate with AIS transmissions across different geographic regions. This analysis employs a straightforward percentage calculation (AIS cross-match count/total count).

High AIS coverage zones, indicated by deep blue coloration, are predominantly concentrated in deeper offshore waters along specific corridors off Karnataka and Kerala, where coverage percentages frequently exceed 75-100%. These high-coverage zones correspond to operational areas likely dominated by larger commercial fishing vessels that are both required to carry AIS equipment and possess the technological capabilities for extended offshore operations.

Conversely, extensive nearshore areas demonstrate critically low AIS coverage percentages, often falling below 25%, particularly in the productive fishing grounds off Maharashtra, Goa, and northern Karnataka. The Goan coastal waters emerge as a particular concern, showing coverage percentages frequently below 10%, indicating that the vast majority of detected fishing activity in these prime fishing zones occurs without AIS transmission.

The spatial distribution of coverage percentages reveals an inverse relationship between fishing activity intensity and surveillance coverage, where the most productive fishing areas demonstrate the lowest AIS coverage rates.

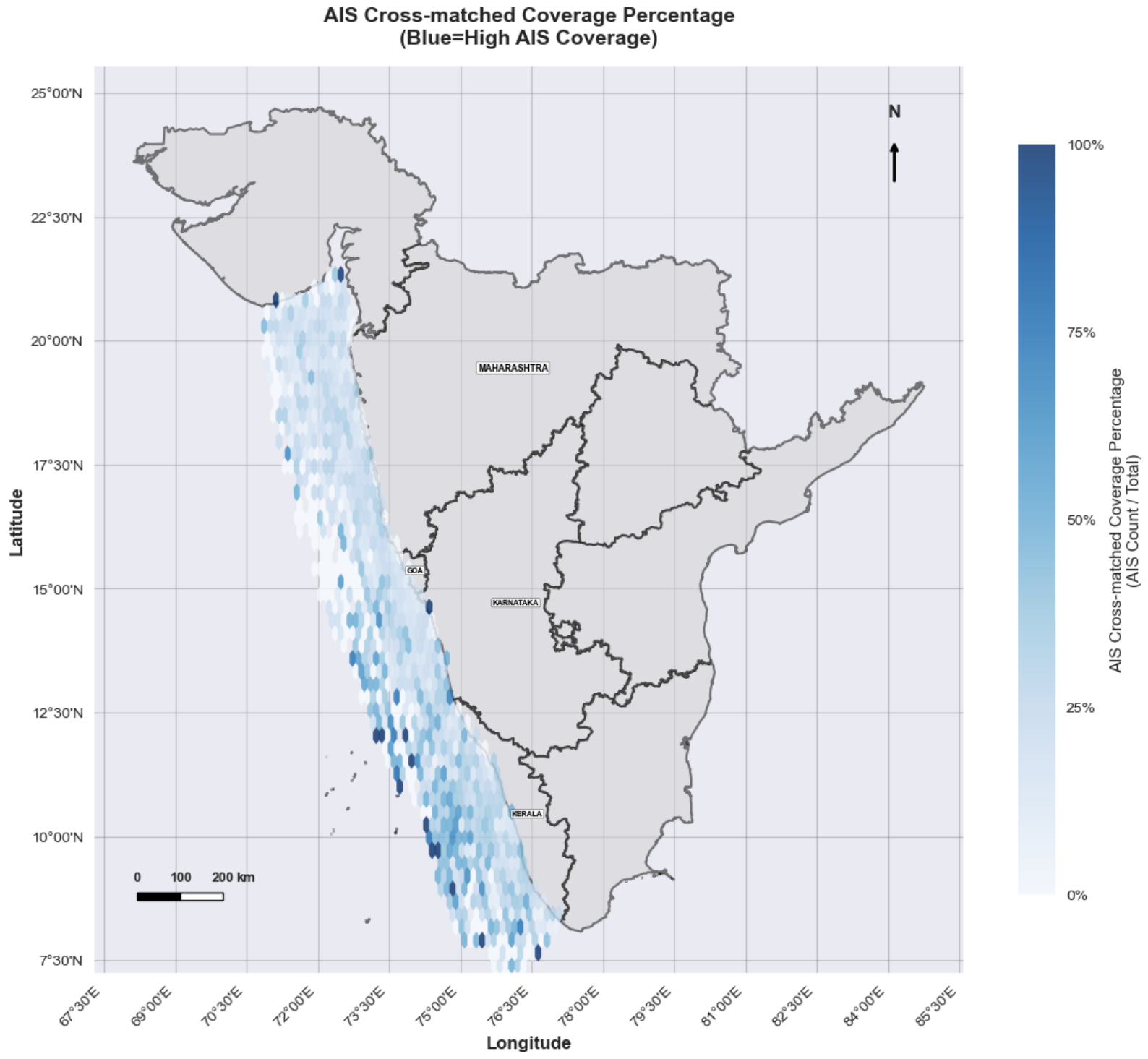

**Figure 13 AIS Cross-matched Coverage Map**

**(ii) Spatial Dominance Analysis (Grid-Level Analysis)**

The spatial dominance analysis employs a normalized dominance index producing values ranging from -1 to +1. The scale interpretation is as follows: -1.0 represents grid cells with AIS-matched vessels only, -0.5 indicates AIS-dominated areas where AIS-matched vessels outnumber dark vessels by 3:1 or greater, 0.0 signifies balanced zones with approximately equal vessel numbers, +0.5 denotes dark vessel-dominated areas where dark vessels outnumber AIS vessels by 3:1 or greater (Hereafter, referred to as Dark Dom), and +1.0 represents grid cells containing exclusively dark vessel detections with no AIS presence (Hereafter, referred to as Dark Only).

The analysis, as shown in **Figure 14** reveals extensive dark vessel-dominated zones concentrated along the nearshore and mid-shelf waters off Maharashtra, Goa, northern Karnataka, and significant portions of Kerala. These zones predominantly fall within the "Dark Dom" and "Dark Only" categories, indicating that 75-100% of detected vessels in these areas might be operating without AIS transmission. The geographic concentration of these high-dominance zones spans approximately 15,000-20,000 square kilometers of prime fishing grounds within the critical 20-100 km offshore corridor.

The persistence of "Dark Only" zones (+1.0 dominance values) across nearshore areas indicates a potential absence of AIS-equipped vessels in regions of intense fishing activity, suggesting either systematic regulatory non-compliance or a fishing industry structure dominated by vessels below AIS mandatory requirements. The Goan coastal waters showcase high dark vessel dominance, with extensive areas showing +0.8 to +1.0 dominance values. AIS cross-matched dominance zones (blue coloration, -0.5 to -1.0 dominance values) appear severely restricted to narrow corridors primarily located in deeper offshore waters along the Karnataka and Kerala coasts.

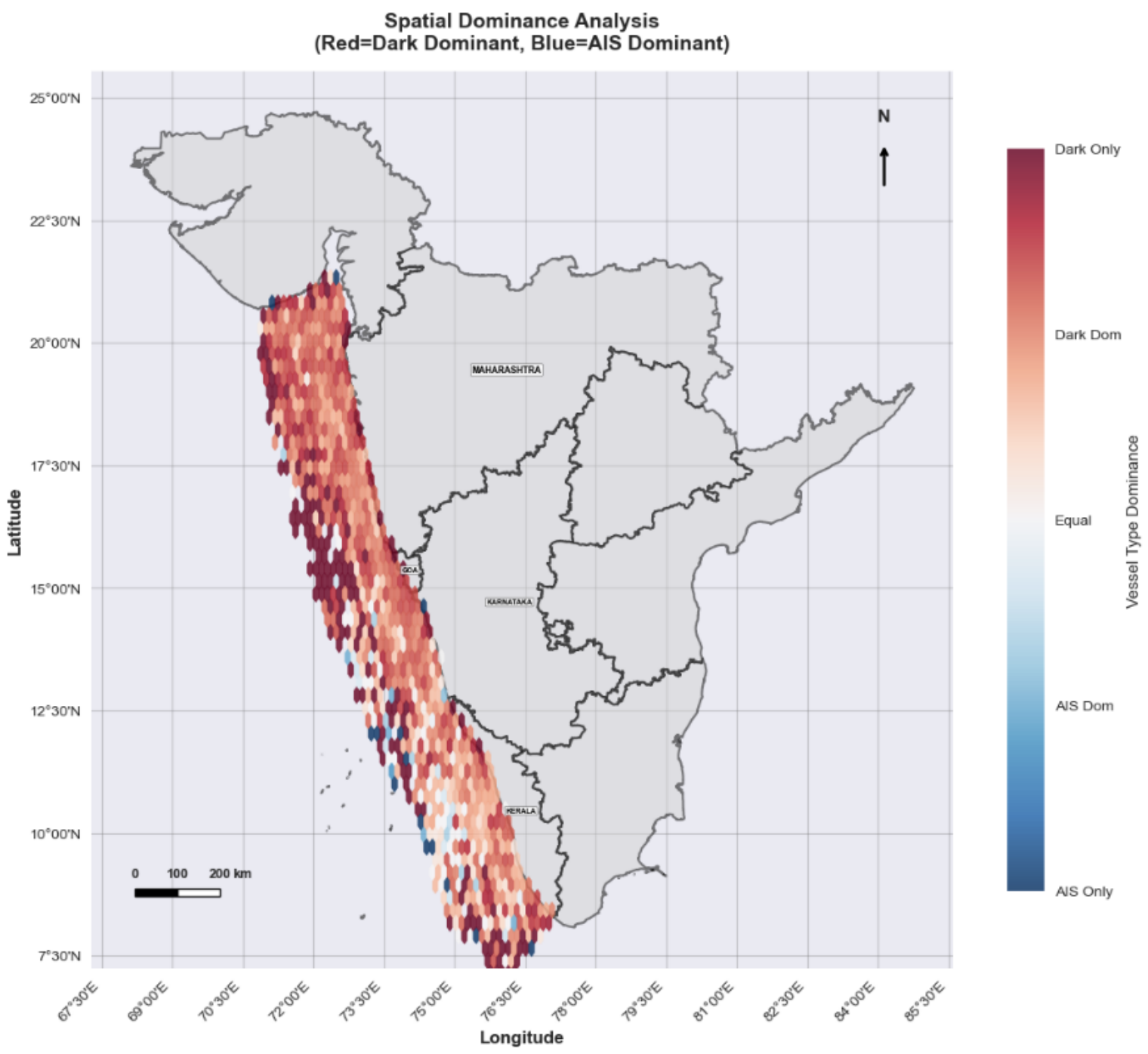

**Figure 14 Spatial Dominance Heatmap**

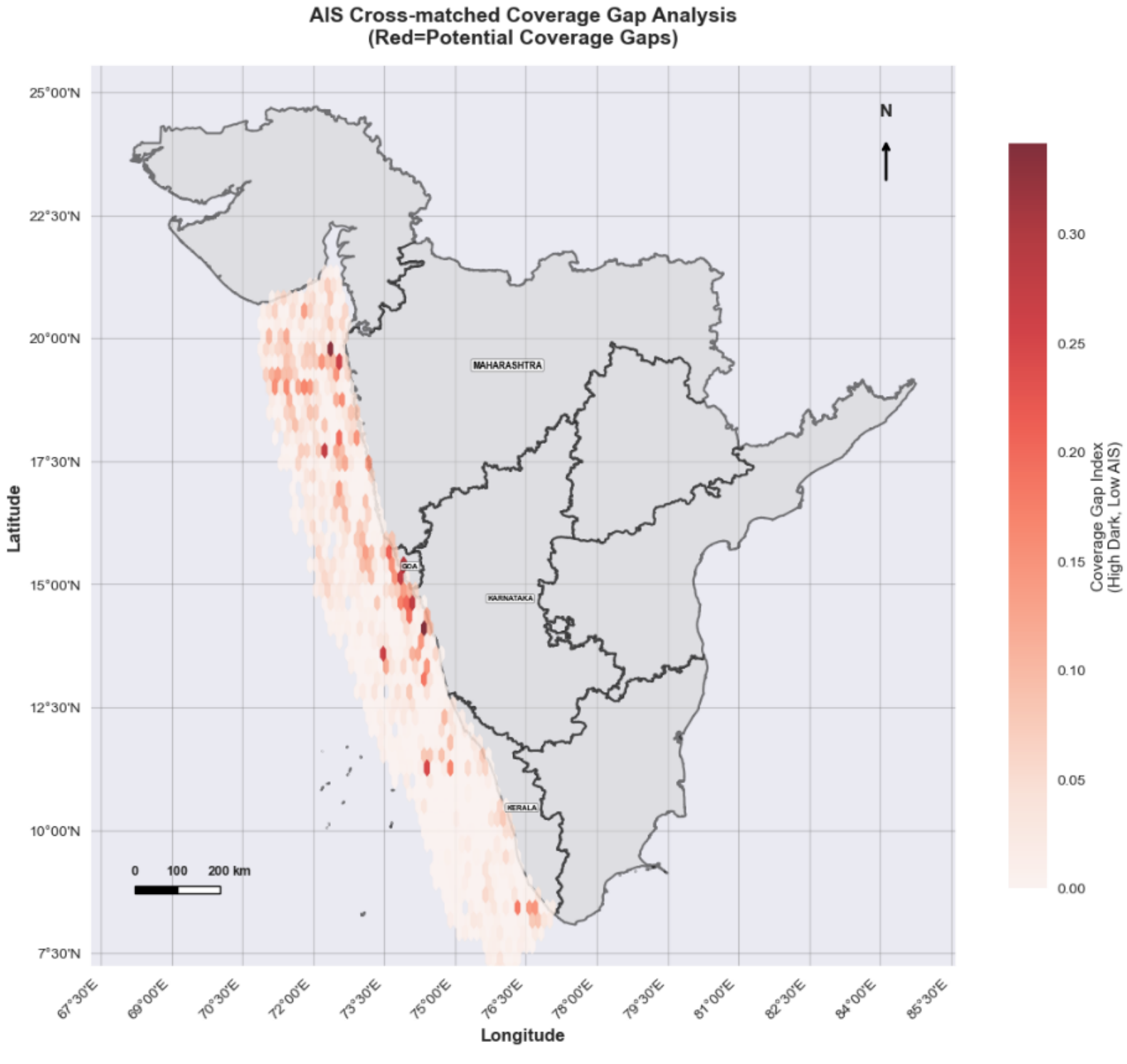

**Figure 15 AIS Coverage Gap Analysis Map**

### (iii) AIS Coverage Gap Analysis (Global Analysis)

The analysis reveals that high AIS coverage zones (75-100%, represented by dark blue coloration) are severely limited to narrow offshore corridors primarily concentrated beyond the 100-kilometer mark from the coastline. As seen in **Figure 15** these high-coverage areas appear as isolated patches along the deeper waters off Karnataka and southern Kerala, representing less than 10-15% of the total fishing activity area. The concentration of high coverage in these remote offshore zones might indicate that AIS-based surveillance systems effectively monitor only the operations of larger commercial vessels capable of extended offshore fishing expeditions, while remaining largely ineffective for nearshore fishing activities.

The nearshore waters off Maharashtra, Goa, northern Karnataka, and Kerala consistently demonstrate coverage percentages below 25%, with vast areas showing coverage rates of less than 10%. The Goan coastal waters continue to present a potential surveillance vulnerability, displaying coverage percentages frequently approaching zero despite intense fishing activity visible in the nighttime light detections.

## 4. Discussion

This section discusses the key takeaways from the study, examining the technical innovations, performance achievements, and broader implications of the SDGSAT-1-based vessel detection methodology developed in this research. The discussion provides a critical analysis of the dual-branch YOLO11 architecture, evaluates model performance against existing approaches, and assesses the practical implications for maritime surveillance applications.

### 4.1 Architectural Improvements and Model Performance

| Model Type | Precision | Recall | F1 Score | mAP@50 | mAP@50-95 |
|---|---|---|---|---|---|
| YOLOv5s | 0.92 | 0.85 | 0.88 | 0.92 | 0.41 |
| YOLOv8s | 0.81 | 0.94 | 0.87 | 0.94 | 0.45 |
| YOLO11s | 0.91 | 0.96 | 0.93 | 0.96 | 0.44 |
| **Dual-Branch YOLO11** | **0.99** | **0.94** | **0.96** | **0.96** | **0.46** |

**Table 5 Model Comparison Results**

The YOLOv5s and V8s have been used in several previous studies of vessel detection using NTL imagery(Hu et al., 2024; Shao et al., 2021; Song et al., 2023; Zhong et al., 2020)**.** The results in **Table 5** clearly show that the dual-branch YOLO11 model delivered better performance than the single-branch setups of YOLOv5s, v8s, and 11s across all key evaluation metrics. By combining both PAN and RGB inputs, the dual-branch model was able to make more accurate and reliable detections. Precision and recall both improved, leading to a higher F1 score, which reflects a better balance between false positives and missed detections. Similarly, the boost in mAP@50 and mAP@50-95 indicates that the dual-branch model was more effective at correctly localizing vessels, even at stricter thresholds(B. Zhang et al., 2024)**.** Overall, the integration of both spectral inputs helped the model capture richer information, translating to noticeable gains in detection quality. This improvement aligns with findings from previous dual-branch studies in remote sensing, where multi-modal feature fusion consistently enhances detection accuracy for small targets(Hua et al., 2022; Huang et al., 2022; T. Zhang et al., 2022).

**Metric Stability:** Throughout the training process, all performance metrics demonstrated stable convergence without significant oscillations or instability. The precision and recall metrics showed consistent improvement during the initial training phases and maintained stable values above 0.9 for most of the training period. This stability is crucial for operational deployment, as it indicates predictable and reliable performance.

The absence of significant overfitting, as evidenced by the close tracking between training and validation metrics, suggests that the model architecture and training methodology are well-suited for the fishing vessel detection task. The dual-branch architecture appears to provide sufficient capacity for learning complex vessel features while maintaining generalization capability.

**Performance Consistency:** The model's performance metrics remained consistent across different evaluation phases, with minimal variance in precision, recall, and mAP values. This consistency is particularly important for fishing monitoring applications where reliable performance is required across varying conditions.

## 4.2 Detection Results and Operational Implications

The detection of 31,526 potential fishing vessels from 1.5 million processed patches represents a large enough dataset for fishing activity analysis. This detection density aligns with expected fishing vessel concentrations in the Arabian Sea, where commercial fishing operations are extensive(J. Li et al., 2024).

The average confidence scores ranging from 0.4 to 0.7 suggest moderate to high model certainty in detections. This confidence distribution is particularly valuable for operational applications.

## 4.3 Spatial-Temporal Patterns and Environmental Drivers

The evident seasonality in vessel detections reflects the complex interplay between monsoon patterns, fish migration cycles, and operational constraints along India's west coast. The peak activity during Q1 (January-March) corresponds to the post-monsoon period, when sea conditions are optimal, and fish stocks are thoroughly replenished(J. Li et al., 2024). This pattern aligns with established fishing calendars in the Arabian Sea(Madhupratap et al., 2001).

The pronounced decline in detected fishing activity during the monsoon months (June–August) is primarily driven by two factors. First, India's annual monsoon trawl ban, aimed at conserving marine stocks during their peak spawning season, prohibits all trawling operations along the west coast from 15 June to 31 July (extended in some states through mid-August) to protect breeding fish and allow stock replenishment(Salim, 2007). Second, persistent cloud cover during the Southwest Monsoon severely impedes SDGSAT-1 glimmer sensor acquisitions over coastal waters, resulting in a dearth of usable datasets(X. Li & Hu, 2025).

Spatial concentration patterns reveal distinct fishing grounds and operational preferences. The primary activity corridor parallel to Maharashtra, Goa, and Karnataka coastlines corresponds to productive continental shelf areas with water depths optimal for traditional fishing methods. However, it's seen that the coast of Maharashtra exhibits slightly offshore trends of fishing vessel volumes, which may be attributed to its broader continental shelf compared to the other states in the study area.

## 4.4 Potential Dark Vessels Spatial Distribution

The spatial dominance analysis reveals critical patterns in potential dark vessel activity. The concentration of dark vessels in nearshore waters (20-100 km offshore) represents the prime operational zone for small fishing vessels, where AIS coverage is traditionally weakest.

The AIS coverage gap analysis identifies specific geographic vulnerabilities, particularly along the Goan coast and Maharashtra nearshore areas. The 70-80% coverage gap in high-activity coastal zones highlights the pressing need for enhanced monitoring systems in these areas.

## 4.5 Limitations and Future Scope

The 77.3% dark vessel proportion represents a significant finding with important implications for maritime security and fisheries management. Such a huge disparity between the AIS cross-matched vessels and dark vessels was also observed in a previous AIS cross-matching study by (Hsu et al., 2019).

While this high percentage initially appears concerning, several factors contribute to this observation:

a. Data Limitations: One likely reason for this mismatch between AIS-confirmed vessels and dark detections is the quality and resolution of the AIS dataset used in the analysis. The AIS data was only available at hourly intervals, meaning any vessel that happened to be active during the satellite overpass but didn't transmit within a ±30-minute window could have been missed entirely. This limitation may have led to underreporting of AIS matches and overestimation of dark vessel presence. In short, the accuracy of this cross-matching process is tightly tied to how detailed and frequent the AIS transmissions are. With access to higher-resolution AIS data, ideally with minute-by-minute updates, the detection-to-AIS match rate would likely improve substantially, offering a clearer and more accurate picture of vessel activity in the region.

b. Regulatory Compliance: Indian maritime regulations mandate AIS transmission for vessels over 20 meters in length. However, there might be a possibility that several traditional fishing vessels in the study area operate below this threshold, legally exempting them from AIS requirements.

c. Operational Behaviour: Fishing vessels may legitimately disable AIS transmitters during active fishing to evade detection (Bunwaree, 2023).

d. Spatial Buffer Analysis: The 6.5 km buffer radius, calculated based on average vessel speeds of 7 knots over 30 minutes, represents a reasonable spatial uncertainty zone. However, this buffer size needs refinement based on specific vessel types and operational patterns in different fishing zones. This, in turn, can only be possible with a more pronounced AIS dataset.

Overall, the study faced challenges regarding AIS data quality; the hourly resolution of available AIS data likely contributed to cross-matching errors, potentially overestimating the dark vessel count. Higher-frequency AIS data (minute-level) would refine this analysis. Additionally, SDGSAT-1's optical sensors are limited by cloud cover, resulting in data gaps during the monsoon season. Future integration with SAR data is essential for all-weather monitoring capabilities.

## 5. Conclusion

This study demonstrated the efficacy of high-resolution SDGSAT-1 nighttime light imagery combined with a custom dual-branch YOLO11 deep learning model for maritime surveillance. The research achieved its primary objectives, delivering a model with 0.99 precision and 0.96 mAP@50, significantly outperforming standard architectures. The analysis of 2022-2023 data revealed a potential surveillance gap, with 77.3% of detected vessels operating without AIS transmission. Spatio-temporal analysis identified critical fishing hotspots and seasonal trends governed by the monsoon cycle.

The findings can have potential implications for fisheries management. The identification of "Dark Dominant" zones allows authorities to optimize patrol deployments to areas with high NTL detections but low AIS visibility. The methodology provides a replicable framework for global application, contributing to the sustainable utilization of ocean resources and the fight against IUU fishing. Future work can focus on integrating multi-sensor data (SAR and VMS) to further refine detection capabilities and classification accuracy.

**Author Contributions:** Conceptualization, P.K.G., R.V.M., S.M.; methodology, S.M.; software, S.M.; validation, S.M., P.K.G., R.V.M; formal analysis, S.M.; data curation, S.M.; writing - original draft preparation, S.M.; writing—review and editing, P.K.G., R.V.M., S.M.; visualization, S.M., P.K.G.; supervision, P.K.G., R.V.M.

**Funding:** This research received no external funding.

**Acknowledgments:** This research work was carried out as a part of the M.Sc. dissertation by the first author (SM) as a part of joint education program (JEP) of Faculty ITC, University of Twente, The Netherlands and Indian Institute of Remote Sensing (IIRS), Dehradun. The authors are grateful to the heads of IIRS and Faculty ITC, University of Twente for the necessary facilities, support and encouragement.